# Token Optimization and Context Window Management in Multi-Agent AI Workflows

**Author:** Dvir Shamay

*Independent Research*
**Companion file:** `TOKEN-OPTIMIZATION-TECHNICAL-APPENDIX.pdf` — technical appendix with full evidence tables and robustness detail.

---

## Abstract

Multi-agent AI workflows are increasingly limited not only by model quality but by token cost, latency, and context-window quality. Existing work describes agent orchestration patterns, memory architectures, and model-level optimizations, but gives practitioners limited evidence about how to reduce token usage across a full production workflow without damaging output quality.

This paper presents a practitioner framework for token optimization and context-window management in multi-agent AI systems. The framework is grounded in an internal production dashboard that ingests meetings, email, and chat content, extracts structured work items with LLMs, and routes summaries across multiple workstreams. Six optimization patterns are described: context stratification, fetch-once/process-locally architecture, schema-contracted prompts, token-aware fallback chains, semantic caching, and inter-agent communication compression. In the production pipeline, these patterns reduced measured cold-load latency to 61-116 seconds (six timed runs) from an operational pre-optimization baseline of roughly 3.5-10.5 minutes, and historical token analysis estimates a 60-70% token reduction.

The paper also reports a controlled context-composition study: 2,420 confirmatory trials across 11 model configurations, using 661 anonymized workplace communication items scored for relevance. The central result is counter-intuitive. Holding the prompt at a fixed ten items, replacing some high-relevance items with same-domain low-relevance items improves the model's relevance-score concordance on the target items, compared with providing only high-relevance items. We call this **relevance-contrast context**. A candidate explanation, not demonstrated here, is that low-relevance examples help the model calibrate what counts as relevant. In the all-11 paired analysis, the 50:50 signal/noise condition improved relevance accuracy by +0.077 over the 100% signal condition (naive 95% CI [+0.056, +0.098], Cohen's d

= 0.49, Holm-adjusted $p < .001$, $n = 220$). These 220 configuration-by-block cells are not independent; summarized by the nine model families the effect is +0.084 (95% interval [+0.064, +0.103]), and we report it as a within-corpus descriptive comparison rather than a population inference. A follow-up Fusion-of-N study found that learned synthesis of repeated extraction samples did not beat the mechanical set union of extracted item IDs, suggesting a simple deployment rule for recall-oriented extraction.

The contribution is a measured engineering layer between model research and production agent practice: not a new model architecture, but a set of repeatable patterns and evaluation methods for making agent workflows faster, cheaper, and more reliable.

---

## 1. Introduction

Agentic AI systems often trade cost and latency for improved task performance. A coordinator calls multiple LLM-powered steps, each step receives context, each output becomes context for a later step, and fallback or review loops may repeat work. This makes token usage compound quickly. A workflow that appears modest from the user's point of view can consume tens or hundreds of thousands of tokens internally.

The cost problem is not only financial. Large or poorly arranged context windows can degrade quality by burying useful evidence among irrelevant material. Developers therefore face a practical question that is still under-specified in the literature: **what should be sent to the model, in what form, and at what stage of the workflow?**

This paper addresses that question at the orchestration layer. It does not claim to improve model internals, invent agent workflow taxonomies, or replace existing memory architectures. Instead, it asks how builders can make existing agent workflows efficient and measurable.

### 1.1 Contributions

This paper makes four contributions.

1. It defines six token-optimization patterns for multi-step LLM workflows, each tied to a production implementation and an observed cost, latency, or quality effect.
2. It frames **context-window management** as an engineering discipline: a set of skills for pruning, compressing, routing, and measuring context.

3. It reports a controlled context-composition experiment across 11 model configurations and 2,420 confirmatory trials, showing that same-domain low-relevance context can improve relevance-score concordance on target items by providing relevance contrast.
4. It reports a focused Fusion-of-N follow-up showing that, for recall-oriented extraction, the set union of repeated samples is a strong mechanical baseline that learned fusion did not beat in this setting.

This paper does not invent new agent-orchestration patterns, agent memory architectures, or model-level prompt caching, and its claims are scoped accordingly. It does not assert that same-domain noise helps every task, nor that learned fusion is useless in general. Its contribution is the practitioner layer described above. That layer is an end-to-end account of what changes token and latency cost in a real multi-stage workflow, and a teachable framing for the engineering discipline behind those changes.

---

## 2. System and Optimization Patterns

The system under study is an internal production dashboard for tracking multiple concurrent engineering workstreams. It ingests recent meetings, email threads, and chat threads, extracts structured information, matches extracted items to workstreams, and generates status summaries.

The optimized architecture uses three stages: data acquisition, extraction, and assignment/summarization.

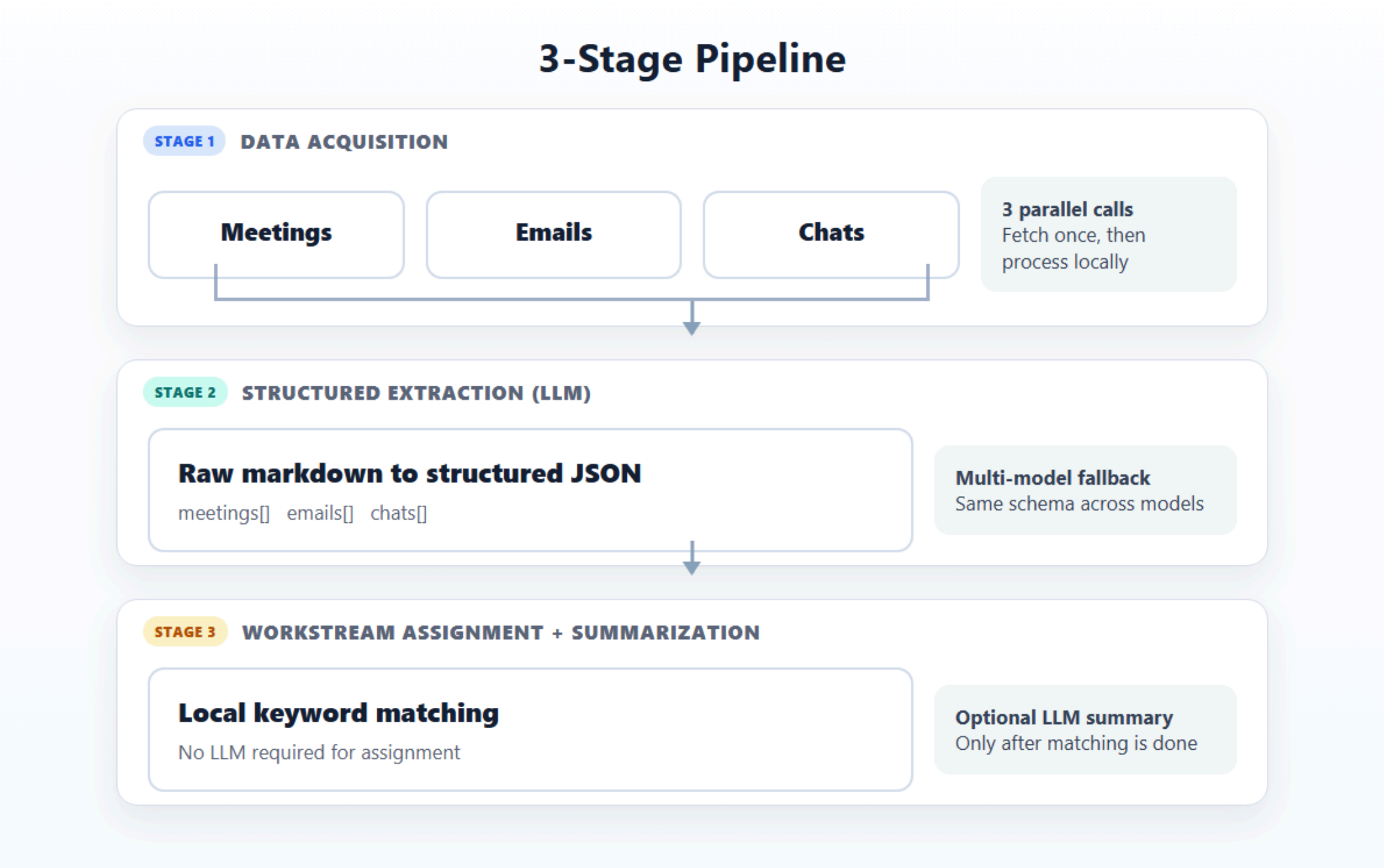

Figure 1. Three-stage pipeline architecture.

## 2.1 Six Optimization Patterns

The six patterns are summarized below. Detailed implementation notes, aggregate pipeline run summaries, and telemetry specifications are kept in the technical appendix rather than the publication body.

| Pattern | Problem Addressed | Implementation | Observed Effect |
|---|---|---|---|
| Context stratification | Models receive all context for all subtasks | Route only the relevant data type to each extraction call | Input per extraction call reduced from roughly 30K tokens to roughly 3K tokens (estimated) |
| Fetch once, process locally | Sub-agents repeatedly fetch the same source data | Acquire meetings/email/chats once, then perform local matching | Data-acquisition API calls per run cut from 7-21 down to 3, and cold-load latency cut to 61-116 seconds |
| Schema-contracted prompts | Models spend output tokens on prose, caveats, and formatting | Require compact JSON contracts and explicit empty-result conventions | Lower output token usage and higher parse reliability (qualitative) |
| Token-aware fallback chains | Rate limits cause full pipeline retries | Retry extraction with validated fallback models without rerunning acquisition | Higher availability without repeating the costly data-acquisition stage, the multi-call fetch that dominates cold-load latency |
| Semantic caching | Repeat loads resend identical context | Cache full pipeline results and validate against current request parameters | Cached repeat loads skip re-fetch and re-tokenization with zero LLM tokens (near-instant in operation; not benchmarked) |
| Inter-agent communication compression | Agents pass verbose human-readable status messages to other agents | Replace prose handoffs with schema-contracted machine messages | Estimated 40-60% reduction in wrapper tokens for inter-stage messages |

The unifying principle is simple: **send less context, but send context with better structure and better timing.** The system saves tokens not by hiding information from the model, but by ensuring each model call receives only the information needed for that step.

## 2.2 Context-Window Management as an Engineering Skill

The optimization patterns point to a broader discipline: context-window management. Just as memory management and state management became first-class software engineering concerns in earlier eras, context-window management is becoming central to AI workflow engineering.

The core skills are:

1. **Context pruning:** knowing when old conversation or task state has become harmful.
2. **Information density:** conveying intent with fewer tokens and less ambiguity.
3. **Memory offloading:** moving stable facts into files, databases, or structured memory rather than repeatedly placing them in the prompt.

4. **Parallel discovery:** collecting independent context sources concurrently to reduce wall-clock latency.
5. **Context pollution detection:** recognizing when irrelevant context is changing model behavior.
6. **Context calibration:** finding the task-specific balance between too little context and too much context.

The empirical study in Section 4 refines this calibration idea. For extraction tasks, the danger is not always "too much noise." Same-domain low-relevance items can help the model set a relevance threshold.

---

## 3. Measurement Design

The evaluation has two parts: a production pipeline analysis and a controlled context-composition experiment.

### 3.1 Production Pipeline Analysis

The production analysis uses 25 recorded pipeline runs from February 2026. Six of the runs have measured elapsed times, and the remaining runs were captured before full telemetry was deployed. Because token usage was not instrumented from the beginning, historical token counts are estimated using character-count approximations and later validation against API usage metadata. This is an important limitation, and the paper treats token reductions as estimates rather than exact measurements.

The production comparison is nevertheless useful because the optimized cold-load latency is directly observed from recorded runs at 61-116 seconds. The pre-optimization baseline of 3.5-10.5 minutes is an operational estimate captured before full telemetry was deployed, and cache-hit loads are near-instant in operation but were not separately benchmarked.

### 3.2 Context-Composition Experiment

The controlled study uses 661 anonymized communication items from one knowledge worker's participant-centered communication stream over a single work week: 15 meeting transcripts, 627 email threads, and 19 chat threads. Each item was scored by the author on a 0-3 relevance scale: 3 marks a decision, action item, blocker, or commitment that affects current work; 2 marks relevant status or context a teammate would want in a weekly summary, but not action-critical; 1 marks a tangential mention with no decision, action, or notable status; 0 marks an item with no bearing on active engineering work (the full rubric is in Appendix D). Items scored 2 or 3 are treated as signal; score-0 items are treated as low-relevance same-domain context. Score-1 items are excluded from the confirmatory experiment to sharpen the contrast.

The confirmatory experiment includes 2,420 trials across 11 model configurations. Every model configuration ran 20 trial blocks per condition. Each prompt contains 10 items, and configurations vary either the signal/noise ratio or the position of signal items within the prompt.

The ratio configurations compare 100% signal, 80% signal, 50% signal, 30% signal, and 10% signal (that is, 10, 8, 5, 3, and 1 signal items out of the 10). The six position configurations hold the signal/noise ratio at 50:50 and place signal items at the start, middle, or end, or scatter them through the prompt in one of three scatter patterns.

The primary metric is **Relevance Accuracy (RelAcc)**: agreement between the model's 0-3 relevance score and the human annotation, giving full credit for an exact match and partial credit when the scores are adjacent. Writing *h* and *m* for the human and model 0-3 scores of an item and *T* for the set of target (signal) items:

$$\mathrm{RelAcc} = \frac{1}{|T|}\sum_{i \in T} s(h_i,\, m_i)$$

$$s(h,m) = \begin{cases} 1.0 & |h-m| = 0 \\ 0.5 & |h-m| = 1 \\ 0.0 & |h-m| \geq 2 \end{cases}$$

Unparseable model output scores 0. Because RelAcc is computed over signal items, it measures relevance scoring on target items rather than full extraction precision. Fact-type recall, precision, and JSON parse success are tracked as secondary metrics. Fact-type recall is, for each target item, the fraction of the human-annotated fact types the model recovered, with partial credit for near-equivalent types (for example, a decision reported as an action); reported values are means of that item-level score.

The contrasts below were specified in a pre-run design-review prompt (not a registry-backed preregistration; that prompt covered a smaller model set, and the final analysis set was elaborated from it before scoring). They are evaluated as paired comparisons over model x trial-block pairs, with Holm correction for multiple testing.

### 3.3 Fusion-of-N Follow-up

The Fusion-of-N follow-up asks whether repeated extraction samples should be merged by another LLM or by a simpler mechanical rule. A single model generated five independent extraction samples per window. The study compared a single sample, best-of-N judging, LLM fusion, and the set union of extracted item IDs.

This study is narrower than the main context-composition experiment. It uses one generator/judge model (DeepSeek-V3.2, run at its default setting), one corpus, and a self-fusion setup where the generator and fuser are the same model. Its role in this paper is practical rather than universal. It tests a common deployment intuition that "an LLM should merge multiple samples intelligently."

---

## 4. Results

### 4.1 Production Pipeline Impact

The production pipeline results show large operational gains from orchestration-layer optimization.

| Metric | Before Optimization | After Optimization | Notes |
|---|---|---|---|
| Cold load latency | 3.5-10.5 min (operational estimate) | 61-116 sec | Optimized value observed from recorded runs; pre-optimization baseline is an operational estimate |
| Cached repeat load | Not available | Near-instant | Full pipeline cache hit; no LLM calls; operational, not benchmarked |
| Estimated input tokens | 50K-80K per run | 60-70% lower | Historical estimate, not exact telemetry |
| Data acquisition calls | 7-21 | 3 | Fetch once, then process locally (operational count) |
| Parse success | Variable before schema contracts | Higher after schema contracts | Qualitative operational observation; parse-success rate was not separately instrumented before and after |

These results should be interpreted as a production case study rather than a benchmark. The exact latency values depend on the source systems and model endpoints. The transferable result is the pattern: eliminate redundant acquisition, reduce prompt surface area, and cache validated pipeline outputs.

### 4.2 Relevance-Contrast Context: Low-Relevance Same-Domain Items Help

The central empirical finding is that, for relevance scoring on target items, a context window containing only high-relevance items is not optimal. Models achieve higher relevance-score concordance when the prompt includes same-domain low-relevance items that demonstrate what “not important” looks like. This is a target-only concordance result, not a demonstrated gain in overall extraction precision.

| Configuration | RelAcc | Fact-type recall |
|---|---|---|
| 100% signal | 0.584 | 0.499 |
| 80% signal | 0.595 | 0.499 |
| 50% signal | 0.661 | 0.522 |
| 30% signal | 0.661 | 0.542 |
| 10% signal | 0.648 | 0.545 |

Each RelAcc and recall value in the table is the mean across all 11 model configurations and 20 trial blocks for that condition. Relevance accuracy rises from the pure-signal condition to a broad plateau. The 50:50 and 30:70 mixes are tied (0.661) and the 10:90 mix is only marginally lower (0.648), so the benefit is best read as spanning roughly 30-50% signal (50-70% same-domain noise) rather than sitting at a single optimum, with fact-type recall continuing to rise as the signal share falls.

The primary contrast, 50% signal versus 100% signal, improves RelAcc by +0.077 (95% CI [+0.056, +0.098], Cohen's d = 0.49, Holm-adjusted $p < .001$, n = 220 model-configuration x block pairs). All 11 model configurations show a positive raw direction, though the effect is negligible for the maximum-reasoning variant discussed below. Because the ratio conditions use nested signal subsets, we also ran a matched-item sensitivity analysis that scores the 100% signal condition only on the same five signal items present in the 50:50 condition, and the effect remains +0.076 (95% CI [+0.053, +0.099], $p < .001$). Because model configurations are not independent, we also summarize the effect with the model family as the unit of analysis. The eleven configurations reduce to nine model families. The three Claude Opus 4.7 reasoning settings count as one family, while GPT-4o, GPT-5.4, and GPT-5.5 are separate families (GPT-5.5's high- and maximum-reasoning runs shown in Figure 2 were added post hoc and lie outside this eleven-configuration set). All configurations also share the same twenty trial blocks. Averaging the nine per-family effects gives +0.084 (95% interval [+0.064, +0.103] across nine families), and clustering by trial block gives [+0.032, +0.122]. Both intervals exclude zero. With only nine families, we treat this as a bounded finite-corpus descriptive effect rather than a population-level inference. The full model list is in Appendix D.

As an additional robustness check, we re-ran the confirmatory ratio comparison as a separate two-arm replicate — the original corpus versus a corrected corpus (one exact-duplicate item and one out-of-scope item removed, and export metadata such as timestamps and header lines stripped from the affected chat items) — while holding the original human relevance labels fixed. This replicate covers ten of the eleven model configurations (Llama-3.3-70B was retired at rerun time), so its original-corpus arm (+0.092) is a fresh measurement, not the frozen +0.077 headline. The gain stayed clearly positive on both arms — +0.092 original and +0.083 corrected — both in the range of the +0.077 headline. The small corrected-minus-original difference (−0.008) did not survive sensitivity checks and is not read as a real reduction of the effect (Section 7 and Appendix D give the full two-arm result).

This result does not mean arbitrary irrelevant context is helpful. The low-relevance items are same-domain workplace communications, not unrelated distractors. The likely mechanism is threshold-setting. When all items are relevant, the model has no examples of ordinary low-priority content and tends to over-rate relevance. Because the primary RelAcc metric is signal-item focused, this mechanism is a hypothesis supported by relevance-score behavior, not a complete precision/F1 explanation.

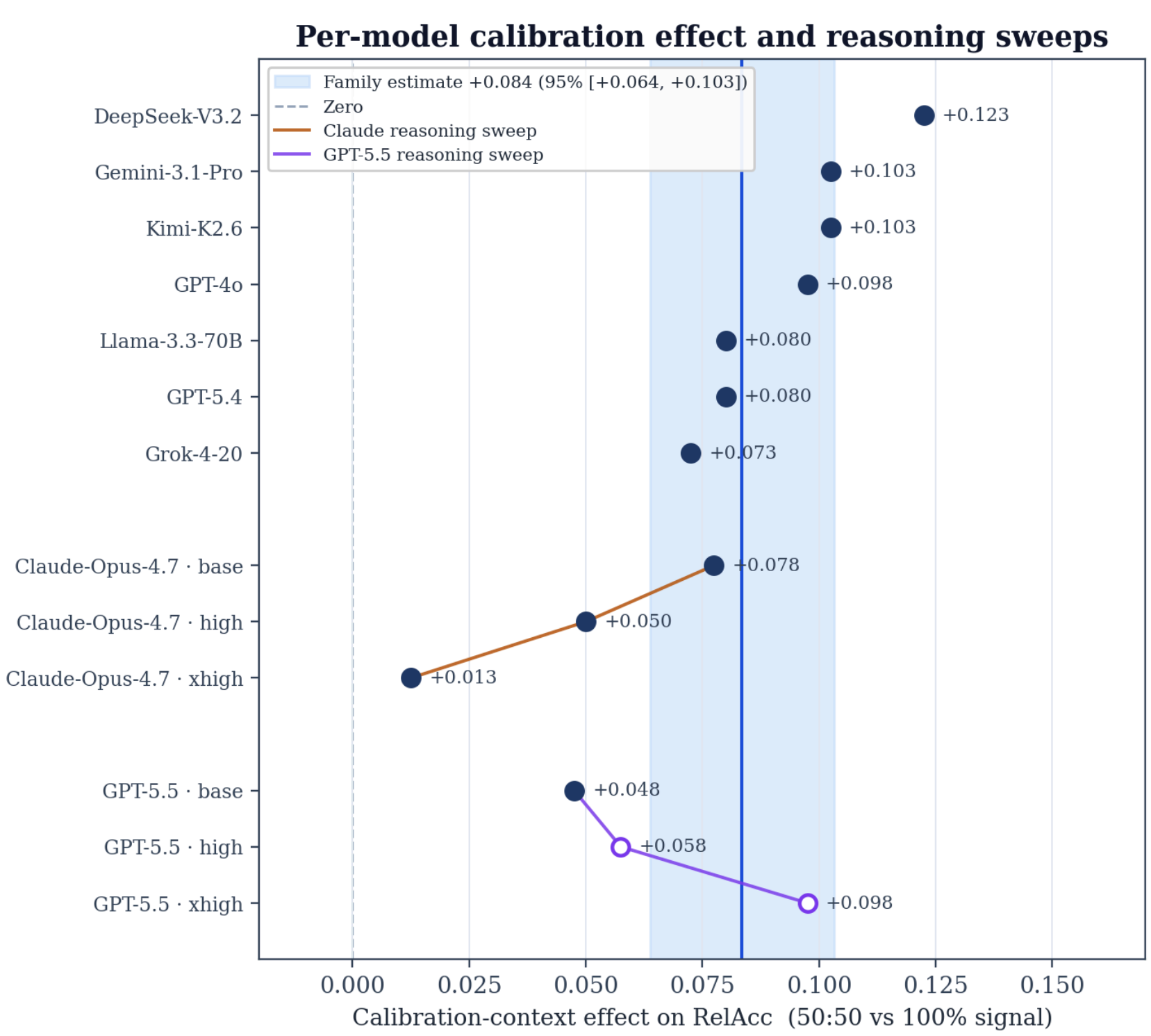

Hollow markers = post-hoc GPT-5.5 reasoning runs (outside the eleven-configuration set). The two sweeps' signed divergence is exploratory (see paper Section 4.5).

Figure 2. How much each model's target-item scoring improves when same-domain low-relevance items are added to the prompt. Each point is a configuration's RelAcc gain for the 50:50 versus 100% signal condition. The solid line marks the model-family estimate (+0.084) and the shaded band its 95% interval, while the dashed line marks zero. The effect is positive for every configuration. Connected points trace the two reasoning-budget sweeps (Claude-Opus-4.7 and GPT-5.5), and hollow markers are the post-hoc GPT-5.5 reasoning runs outside the eleven-configuration set. The two sweeps move in opposite directions, and this divergence is exploratory (Section 4.5).

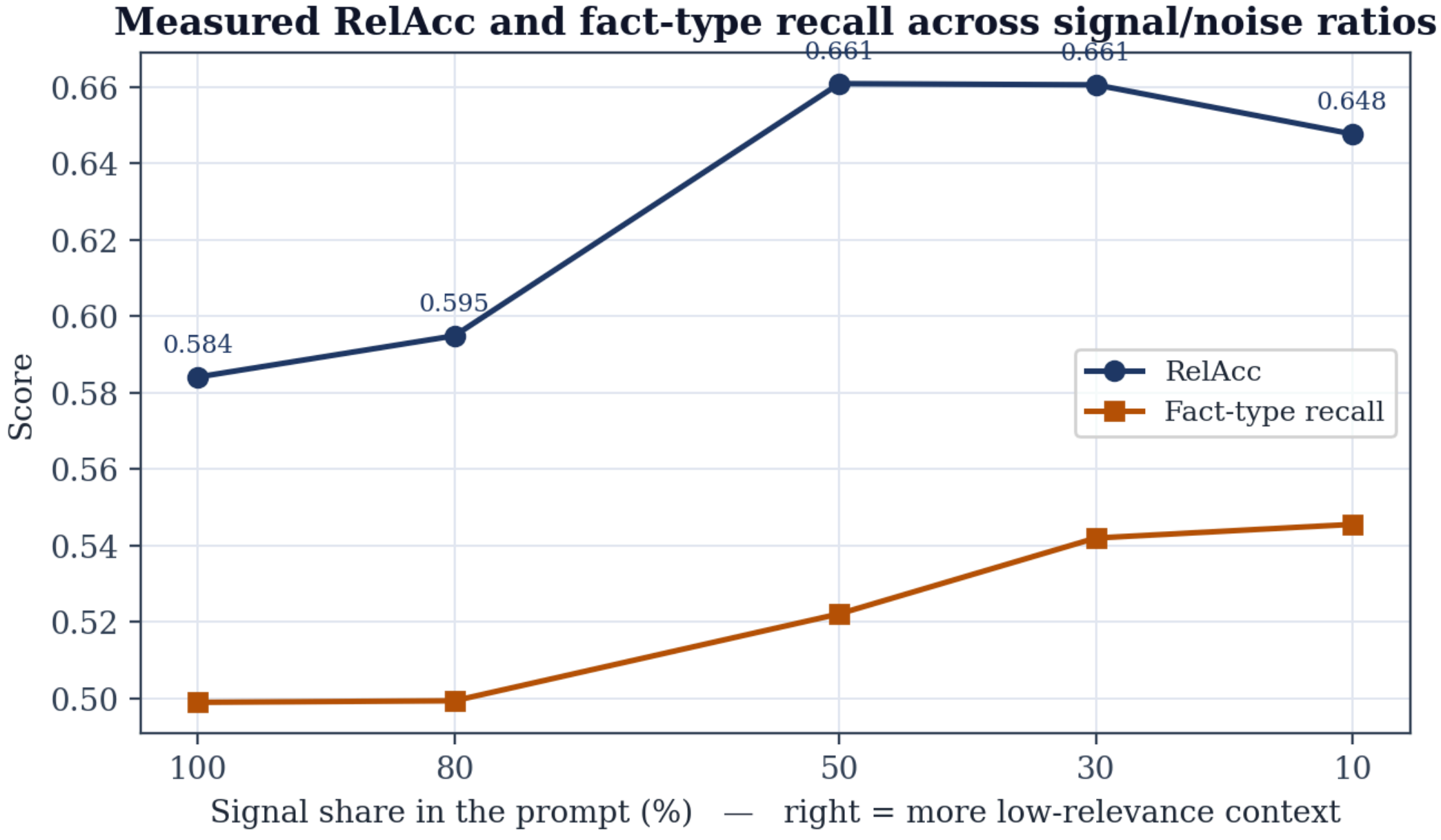

Figure 3. Two quality measures as the prompt shifts from all-signal to more same-domain noise. Relevance accuracy (how closely the model's 0-3 target-item scores match the human labels) is highest across a broad plateau, with the 50:50 and 30:70 mixes tied, rather than at a single point, while fact-type recall (the mean share of each target item's human-annotated fact types the model recovered) keeps rising as the signal share falls.

### 4.3 Position Effects Are Smaller and Model-Dependent

The position experiment finds a smaller effect than the ratio experiment. Across all 11 model configurations, placing signal items at the end of a 10-item prompt improves RelAcc by +0.022 over placing them at the start ($d = 0.14$, unadjusted $p = .033$), but this effect does not survive Holm correction across the five planned contrasts ($p_{adj} = .098$). The expected "lost in the middle" effect is not detected at this scale. Scattering signal items through the prompt did not beat contiguous placement either (contrast C3, null). Middle placement is statistically indistinguishable from the average of start and end placement, with low power for small middle-placement penalties.

This does not contradict prior long-context findings. The prompt windows here are short by long-context standards: 10 items, roughly 2,000-3,000 tokens. A separate exploratory condition with 30-item windows suggests that burial effects emerge at larger prompt sizes. The practical rule is therefore scale-dependent. For short extraction windows, ratio matters more than position. For longer retrieval windows, do not bury the most important content in the middle.

### 4.4 Domain Filtering Is a Large Quality Lever

The exploratory phase finds that processing each communication type separately can be better than mixing meetings, emails, and chats in one prompt when source types differ sharply in signal density. In this dataset, an exploratory (non-confirmatory) analysis found that routing to meeting-only content raised RelAcc by roughly +0.16 (16 points on a 0-100 scale) over mixed-source prompts. Because this comparison is exploratory, it was not part of the pre-specified confirmatory experiment, and we report it as indicative rather than confirmed.

The reason is compositional rather than semantic. In this dataset, meetings are naturally high-signal, with 73.3% of meeting items relevant, compared with 5.1% of email items. Routing by content type therefore changes the signal distribution before the model sees the prompt. This supports Pattern 1. Context stratification is not only a token-saving technique, but also a quality-improvement technique.

### 4.5 Reasoning Budget and the Calibration Effect (Exploratory)

The study includes a full reasoning-budget sweep for one model family (Claude-Opus-4.7, in the confirmatory set) and, added afterward, a matching sweep for a second (GPT-5.5). The point estimates move in opposite directions. For Claude the calibration-context gain falls as the reasoning budget rises (+0.078 default, +0.050 high, +0.013 maximum), as if extended reasoning let the model infer the relevance threshold internally rather than needing examples of low-relevance items. For GPT-5.5 the gain rises with reasoning (+0.048 default, +0.058 high, +0.098 maximum).

Taken at face value this mirror image is striking, but we do not treat it as a real effect. Neither within-model trend is statistically distinguishable from noise (paired by trial block, $p = 0.10$ for Claude and $p = 0.15$ for GPT-5.5). The signed difference between the two families — the interaction that the “opposite directions” reading depends on — is nominally significant ($p \approx 0.015$, $n = 20$ blocks), but it rests on only two model families, the GPT-5.5 sweep was added post hoc, and the test is not corrected for the several reasoning contrasts we examined — so we report the opposite-direction pattern as an exploratory hypothesis, not a finding. Establishing whether reasoning budget modulates the calibration effect would require a dedicated, adequately powered sweep across many model families. Figure 2 shows every configuration, including both reasoning sweeps, and the per-configuration numbers are tabulated in Appendix D.

### 4.6 Fusion-of-N: The Set Union Is a Strong Baseline

The Fusion-of-N follow-up finds that learned LLM fusion does not beat the mechanical union of repeated extraction samples in this setting. The set union of item IDs reaches the observed recall ceiling, and the fuser tends to drop lower-support true positives that only one sample caught.

The prompt-invariant conclusion is not "fusion is always bad." A recall-leaning fuser prompt improves fusion and changes the comparison against best-of-N. The robust conclusion is narrower and more useful. For recall-oriented extraction, if multiple samples identify item IDs, the set union is a simple, strong baseline, and in this study learned fusion did not beat it, so its extra compute was not justified here.

---

## 5. Practitioner Guidance

The results translate into five deployment rules.

1. **Do not over-filter extraction prompts.** If the task is to identify important items, include some same-domain low-relevance items so the model can infer the relevance threshold. In this corpus the gain concentrated in the low-base-rate source (email) and was near zero for meetings and chats, so test it per source type before relying on it.
2. **Route by content type before calling the model when source types differ in signal density.** Meetings, emails, and chats should usually be extracted separately, but the gain depends on the source buckets having different signal/noise profiles.
3. **Optimize context arrangement before upgrading models.** Arrangement changes are cheap to test and, in this study, produced measurable quality changes, so they are worth trying before committing to a model upgrade.
4. **Treat position as a secondary lever.** At short context sizes, ratio matters more. At longer context sizes, avoid burying important content in the middle until that assumption is tested for your task.
5. **Benchmark learned fusion against set union.** If repeated extraction samples are available, union the item IDs first, and only add a fuser if it beats that mechanical baseline on the target metric.

These rules are intentionally operational. The goal is not to produce a universal law of prompting, but to give builders concrete defaults that can be tested in their own pipelines.

A reference implementation of the arrangement rules above — calibration-ratio batching, type routing, and position ordering — is released alongside this paper as `context_arranger.py` (MIT license, Python standard library only, no external dependencies). It targets practitioners who already have a retrieval or pre-filtering step and want a drop-in post-retrieval arrangement layer. Pass items with an optional relevance hint from your own retrieval score, and it groups and orders them per the rules above. The tool implements the arrangement guidance only. It does not implement or validate a relevance judge. A keyword-based fallback is included for convenience when no precomputed score is available, but that heuristic was not the method used in this study (which used human 0-3 relevance labels) and is not independently validated — replace it with your own retrieval score or a model call before relying on it in production. The tool also supports an explicit position parameter — start, end, or an interleaved scattered mode — a per-model position-preference lookup drawn from the per-model analysis in this paper's Technical Appendix (Appendix D, "Per-model position sensitivity"), a configurable long-context threshold, and optional token-budget-aware batching. See the Technical Appendix for the full per-model breakdown and the tool's own documentation for the complete parameter set and their caveats.

---

## 6. Related Work

Anthropic's "Building Effective Agents" describes common workflow patterns such as chaining, routing, parallelization, orchestrator-workers, and evaluator-optimizer loops. This paper complements that taxonomy by measuring cost and context effects inside those workflows.

Lilian Weng's "LLM Powered Autonomous Agents," Park et al.'s "Generative Agents," and Shinn et al.'s "Reflexion" frame memory and feedback as architectures for long-running agent behavior. This paper focuses on a narrower production question — when memory and context are available, how much of them should be sent to each model call?

Liu et al.'s "Lost in the Middle" shows that long-context models often underuse information placed in the middle of long prompts. The present study finds that this effect is not visible at 10-item extraction scale but may emerge at longer windows, suggesting a scale-dependent boundary.

Shi et al.'s work on distractors shows that irrelevant content can hurt reasoning tasks. Cuconasu et al.'s "The Power of Noise" shows that random documents can sometimes improve RAG question answering. The present study is closest to that second line of work but differs in task and mechanism. It studies relevance scoring and extraction over same-domain workplace communications rather than answer generation over retrieved passages.

Prompt compression pursues the opposite lever to the one studied here. Rather than adding calibration context, it removes low-information tokens so more can fit in a fixed budget. Huiqiang Jiang et al.'s LLMLingua drops tokens that a small language model finds predictable, using a budget controller and iterative token-level compression to preserve meaning at high compression ratios. LongLLMLingua adds question-aware compression and context reordering to counter position bias in long inputs, and LLMLingua-2 (Pan et al.) reframes compression as token classification learned by distillation from a larger model. These methods optimize token cost while holding task accuracy roughly fixed. The present study is complementary and points to a boundary condition. For relevance scoring and extraction, some same-domain low-relevance content is not redundant but load-bearing, because it supplies the contrast a model uses to calibrate what counts as important. This suggests a compression-versus-calibration tradeoff on relevance-sensitive tasks — aggressively pruning "boring but on-topic" content can save tokens while quietly weakening relevance calibration — and characterizing where that tradeoff begins is left to future work.

Provider prompt caching and cached prefixes reduce the cost of repeated prompt prefixes at the model-service layer. The semantic caching pattern in this paper operates at the workflow layer. It avoids the model call entirely when the request semantics match a previously computed pipeline result.

Li et al.'s "More Agents Is All You Need," self-consistency, Mixture-of-Agents, and LLM-Blender motivate sampling and aggregation. The Fusion-of-N result in this paper adds a deployment caution. For extraction recall, a mechanical set union may be hard for learned fusion to beat.

Positioned against this literature, the paper's contribution is a practitioner engineering layer rather than a new architecture or a single benchmark: an end-to-end framework for reducing token and latency cost in multi-stage agent workflows, a teachable framing for context-window management, a controlled context-composition measurement across 11 model configurations (nine model families) and 2,420 confirmatory trials showing that same-domain low-relevance context can improve extraction calibration, and a Fusion-of-N deployment caution showing that, in this recall-oriented setup, a free set union is a strong baseline that learned fusion did not beat.

---

## 7. Limitations, Ethics, and Release Controls

The study has important limitations.

1. The dataset is one knowledge worker's participant-centered communication stream from a single week and domain, and the findings may not generalize to other communication styles, workers, or settings.

2. All human relevance scores were assigned by the author; this is an intentional single-annotator design. Because the labels are held fixed across all conditions (see item 8), the reported contrasts reflect prompt composition rather than label consensus.
3. Historical token counts for the production pipeline are estimated, because full telemetry was added after the initial optimization work.
4. The primary relevance metric is task-specific, signal-item focused, and should not be compared directly to standard benchmark accuracy or precision/recall metrics.
5. The ratio configurations use nested signal subsets. The matched-item sensitivity analysis above reduces but does not eliminate the need for replication with independently sampled item sets.
6. The 220 paired measurements are not independent — they cluster within 9 model families and 20 trial blocks. Accounting for this dependence, the effect is +0.084 with the model family as the unit of analysis (95% interval [+0.064, +0.103]) and [+0.032, +0.122] when clustered by trial block, and both intervals exclude zero. With only nine families, we report this as a bounded finite-corpus descriptive effect and attach no population-level p-value, since generalization to the broader model population would require independently sampled clusters.
7. The loader-visible corpus retains same-domain artifacts — 29 lines of export metadata mixed into chat content, at least one headerless chat, and one exact-duplicate content group — that are reused across trials, all confined to the chat source. A leave-out sensitivity analysis over the existing trials shows these artifacts do not drive the finding. Excluding the affected chat targets raises the calibration-context gain from +0.077 (all sources) to +0.112, and restricting to email targets alone raises it to +0.156, with the effect remaining positive across all 11 model configurations and all 9 model families. Because the effect concentrates in email items while the artifacts sit in the near-zero-effect chat source, the artifacts fall where the effect is weakest rather than driving the reported result, and the corrected-corpus rerun (item 9) tests this directly. This is an exclusion analysis over the original generations rather than a regeneration on a rebuilt corpus, which remains a residual limitation.
8. Because all relevance labels were held constant across both experimental arms, the +0.077 effect reflects a change in prompt composition rather than annotation, so single-annotator subjectivity cannot by itself explain it.
9. A direct re-run on a corrected corpus — one exact-duplicate item and one out-of-scope item removed, export metadata stripped from the affected chat items, and the original labels held fixed — reproduced the gain at about +0.083, with both arms clearly positive. The small negative difference from the original corpus is not defensibly interpretable. It does not survive a strict, symmetric hold-out of the few unparseable prompts, shrinks by about three-quarters under a renormalizing check, and appears even among the version-pinned models that cannot drift between runs, so we report it as descriptive rather than as evidence of a real change. The full two-arm numbers are in Appendix D.

**Ethics.** The corpus is the author's own participant-centered communication stream. Content authored by other participants appears only in anonymized, aggregate form. No raw messages, names, or identifiers are released, and no attempt was made to re-identify any individual.

**Release controls.** The underlying communication data is not published. Raw prompts, raw model outputs, per-item IDs, and message content remain private, and the public paper and appendix report only aggregate statistics, named models, and system descriptions.

The paper should therefore be read as a practitioner systems study with controlled experimental support, not as a universal benchmark or a claim about all people, all models, or all extraction tasks.

---

## 8. Conclusion

Token optimization in multi-agent workflows is an engineering problem, not only a model problem. Production systems can become cheaper, faster, and more reliable by changing how context is acquired, routed, compressed, cached, and measured.

The case study shows that orchestration-layer changes can reduce latency by multiple factors and eliminate repeated token use on cached paths. The controlled experiment adds a more surprising result. For extraction tasks, same-domain low-relevance content can improve relevance accuracy by giving the model contrast. The Fusion-of-N follow-up adds a final deployment lesson. Simple mechanical baselines, such as set union, should be tested before adding another model call.

The broader claim is that context-window management should become a first-class engineering skill. As context windows grow, the question will not be whether we can send more context. It will be whether we can send the right context, in the right structure, at the right time.

---

# Technical Appendix: Token Optimization and Context Window Management

**Author:** Dvir Shamay
*Independent Research*

---

## Purpose and Boundary

This appendix provides the supporting evidence and detail behind the summarized claims in the main paper: aggregate run tables, telemetry design, named-model results, and robustness analysis.

The appendix includes only aggregate or operational metadata. It does not include message bodies, names, email addresses, raw prompts containing private content, raw model outputs, or per-item identifiers.

---

## Appendix A: Historical Pipeline Run Data

This table supports the latency claims in the main paper's production pipeline analysis. It is operational evidence from a live pipeline, not a controlled benchmark. The early runs were not fully instrumented, so only six of the twenty-five logged runs have a measured elapsed time.

| Run | Timestamp | Model | Elapsed (s) |
|---|---|---|---:|
| 1 | 2026-02-11 03:01 | DeepSeek-V3-0324 | 111.9 |
| 4 | 2026-02-16 04:35 | Kimi-K2.5 | 65.3 |
| 12 | 2026-02-16 21:30 | DeepSeek-V3-0324 | 61.4 |
| 22 | 2026-02-17 21:39 | DeepSeek-V3-0324 | 116.0 |
| 23 | 2026-02-17 22:04 | DeepSeek-V3-0324 | 96.9 |
| 24 | 2026-02-18 16:55 | DeepSeek-V3-0324 | 86.7 |

"Default" refers to the production pipeline's default extraction model, DeepSeek-V3-0324 — the head of the fallback chain DeepSeek-V3-0324 → GPT-4o → Phi-4 → Kimi-K2.5 → Llama-3.3-70B → Mistral-Large.

During this window 25 runs were logged, and telemetry captured elapsed time for six. Measured optimized cold-load latency ranged 61.4-116.0 seconds. The pre-optimization baseline of roughly 3.5-10.5 minutes is an operational estimate from before full telemetry, so the speed-up is reported as approximate.

---

## Appendix B: Telemetry Module Specification

The telemetry design records the measurements missing from the early runs.

Key events:

```
pipeline.run.start    -> { runId, options }
pipeline.stage.start  -> { runId, stage, metadata }
pipeline.stage.end    -> { runId, stage, elapsed_ms, results }
pipeline.llm.call     -> { runId, model, input_tokens_est, output_tokens_est }
pipeline.cache.hit    -> { runId, cache_age_ms }
pipeline.fallback     -> { runId, from_model, to_model, reason }
pipeline.run.end      -> { runId, total_elapsed_ms, summary }
```

Recommended future fields:

| Field | Reason |
|---|---|
| `prompt_tokens` | Captures actual input-token cost when provider returns usage |
| `completion_tokens` | Captures actual output-token cost |
| `stage_elapsed_ms` | Separates acquisition, extraction, matching, and summarization latency |
| `cache_key` hash | Enables cache-hit analysis without exposing content |
| `model_family` | Allows cost/quality comparison across model classes |
| `fallback_count` | Measures reliability impact of model routing |

---

## Appendix C: Optimization Impact Summary

| Optimization | Tokens Saved | Latency Saved | Implementation Effort | Evidence Type |
|---|---|---|---|---|
| Context stratification | Roughly 90% per extraction call | Roughly 40% from less data per call | Medium | Operational estimate (pipeline design) |
| Fetch once, process locally | Avoids duplicated raw-data context | 57-86% fewer acquisition calls | Medium | Pipeline architecture and run logs |
| Schema-contracted prompts | Roughly 30% output-token reduction | Lower parse/retry overhead | Low | Prompt comparison and API usage logging |
| Token-aware fallback chains | Avoids full reruns after model failure | Avoids repeating acquisition | Low | Operational design |
| Semantic caching | 100% on cache hits | Near-instant repeat loads (not benchmarked) | Low | Operational run behavior |
| Inter-agent communication compression | Estimated 40-60% wrapper reduction | Lower downstream parsing burden | Low-medium | Prompt/payload analysis |
| Compound effect | Roughly 60-70% total token reduction | Roughly 2-10x faster cold loads vs the 3.5-10.5 min pre-optimization baseline (operational estimate; cached paths near-instant but not benchmarked) | Mixed | Aggregate operational estimate |

The token/latency figures in this table are operational estimates from the production pipeline's design and run logs, not the paper's controlled experiment. The paper's central measured result is the +0.077 relevance-accuracy gain from relevance-contrast context (Section 4.2). The pipeline savings above are a separate, secondary case study.

## Appendix D: Controlled Context-Composition Study Details

### Dataset

| Source Type | Count | Signal Rate |
|---|---:|---:|
| Meetings | 15 | 73.3% |
| Emails | 627 | 5.1% |
| Chat threads | 19 | 63.2% |
| Total | 661 | 55 signal / 595 noise / 11 excluded (score-1) |

Signal is defined as a human relevance score of 2 or 3. Noise is defined as a score of 0. The 11 score-1 (marginal-relevance) items were excluded so the contrast is between clearly relevant (2-3) and clearly irrelevant (0) content. A fuzzy middle would add label noise to the RelAcc metric. 55 signal + 595 noise + 11 excluded = 661.

### Relevance Rubric

All items were scored by the author on a single first-impression pass, using a fixed four-point rubric:

| Score | Label | Definition |
|:---:|---|---|
| 3 | High | Captures a decision, action item, blocker, or commitment that affects current engineering work. |
| 2 | Medium | Relevant status or information a teammate would want in a weekly summary, but not action-critical. |
| 1 | Low | Tangentially related to work: mentions it, but carries no decision, action, or notable status. |
| 0 | None | No bearing on active engineering work (social chatter, newsletters, automated notifications, off-topic). |

### Confirmatory Model Set

The confirmatory experiment used the following eleven model configurations:

| Model | Type |
|---|---|
| GPT-4o | General flagship |
| GPT-5.4 | General next-generation |
| GPT-5.5 | Frontier assistant bridge |
| DeepSeek-V3.2 | General/open-weight style |
| Grok-4-20 | Reasoning |
| Kimi-K2.6 | Reasoning |
| Llama-3.3-70B | Open-weight |
| Gemini-3.1-Pro-Preview | Frontier assistant bridge |
| Claude Opus 4.7 default | General |
| Claude Opus 4.7 high | Extended reasoning |
| Claude Opus 4.7 xhigh | Maximum reasoning |

### Pre-Specified Contrasts

| Contrast | Question | Result |
|---|---|---|
| C1 | Does 50:50 signal/noise beat 100% signal? | Supported, +0.077 RelAcc across all 11 model configurations; matched-item sensitivity +0.076 |
| C2 | Is there a positive trend as noise increases? | Supported, slope +0.000871 RelAcc per % noise |
| C3 | Does scattering signal beat contiguous placement? | Null |
| C4 | Is middle placement worse than start/end? | No middle penalty detected at 10-item scale; low power for small effects |
| C5 | Does end placement beat start placement? | Suggestive only in all-11 analysis: +0.022 RelAcc, unadjusted $p$ = .033, Holm $p_{adj}$ = .098 |

### Robustness, Dependence, and Contamination Sensitivity

All figures below are recomputed from the existing per-trial scores. No additional model runs were performed.

**Direction across units.** The C1 relevance-contrast effect is positive in 11 of 11 model configurations, 9 of 9 model families, and 16 of 20 trial blocks (per-block range -0.132 to +0.236).

**Dependence-aware intervals.** The 220 model-configuration x block pairs are not independent. Treating the model family as the unit of analysis gives +0.084 (95% interval [+0.064, +0.103] across nine families, the three Claude Opus 4.7 reasoning settings counting as one). Clustering by trial block gives [+0.032, +0.122]. Both exclude zero. With only nine families, we report a bounded finite-corpus descriptive effect and no population-level p-value.

**Metric robustness.** The calibration-context effect does not depend on the RelAcc partial-credit weighting. Recomputing the same 50:50-versus-100% contrast under alternative item-level scoring rules keeps it positive: +0.070 under strict exact-match accuracy, +0.083 under an off-by-one tolerance, and +0.059 under fully linear partial credit. Mean absolute error on the 0-3 relevance scale falls by 0.18 points, and every family-clustered interval excludes zero. Across target items the model reproduces the human 0-3 score exactly 39% of the time and within one point 83% of the time; and across all scored items (targets and noise pooled) model and human scores correlate at Spearman +0.48 — so the metric reflects substantive agreement between the model and human scores on these repeatedly-scored items rather than an artifact of the weighting.

**Per-model calibration sensitivity (exploratory).** The C1 gain is positive for every measured configuration, but its magnitude varies about tenfold. The first eleven rows are the pre-specified eleven-configuration set. The last two are post-hoc GPT-5.5 reasoning runs outside it. The two reasoning sweeps (Claude-Opus-4.7 and GPT-5.5) move in opposite directions. The signed difference is only nominally significant and rests on two families, so we report it as an exploratory pattern, not a finding (Section 4.5).

| Model configuration | RelAcc gain (50:50 over 100% signal) |
|---|---:|
| DeepSeek-V3.2 | +0.123 |
| Gemini-3.1-Pro-Preview | +0.103 |
| Kimi-K2.6 | +0.103 |
| GPT-4o | +0.098 |
| GPT-5.4 | +0.080 |
| Llama-3.3-70B | +0.080 |
| Claude Opus 4.7 default | +0.078 |
| Grok-4-20 | +0.073 |
| Claude Opus 4.7 high | +0.050 |
| GPT-5.5 default | +0.048 |
| Claude Opus 4.7 xhigh | +0.013 |
| GPT-5.5 high (post-hoc) | +0.058 |
| GPT-5.5 xhigh (post-hoc) | +0.098 |

**Source concentration.** The gain concentrates in email target items and is near-zero for meetings and chats, consistent with the domain-routing result. Email is the low-base-rate source (5.1% relevant) where same-domain low-relevance contrast most helps calibration.

| Matched signal source | Item-instances | Mean RelAcc gain (50:50 over 100% signal) |
|---|---:|---:|
| Email | 671 | +0.119 |
| Chat | 209 | +0.017 |
| Meeting | 220 | +0.002 |

"Item-instances" counts each scored (model configuration x trial block x matched signal item) triple once — not unique corpus items. The matched-item sensitivity analysis reuses the same 5 signal items per block across all 11 model configurations, so 11 configs x 20 blocks x 5 items = 1,100 instances total (671 + 209 + 220), split here by the source type of the underlying item.

**Contamination leave-out.** The frozen data-quality artifacts (export metadata mixed into chat, one headerless chat, one exact-duplicate group) are confined to the chat source, which also carries the near-zero source effect. Excluding chat target items raises the effect from +0.077 (all sources) to +0.112, and restricting to email targets alone raises it to +0.156 (computed over all email target items; this is not directly comparable to the +0.119 matched-five-item email figure in the source-concentration table above, which uses a different scoring scheme). Both remain positive across all 11 configurations and 9 families, so the artifacts fall where the effect is weakest rather than driving the reported result. This is an exclusion analysis over the original generations; a direct regeneration on a cleaned corpus is reported next.

**Corrected-corpus rerun.** To test the data-quality artifacts directly rather than by exclusion, we re-ran the confirmatory ratio comparison on a corrected corpus. The corrected corpus removes one exact-duplicate item and one out-of-scope item (661 → 659 items, 55 → 53 signal) and strips the export metadata (timestamps and header lines) that had leaked into the affected chat items. The original human relevance labels are held fixed, so only the prompt inputs change. The rerun is a two-arm comparison — the original corpus versus the corrected corpus — over a ten-model rerun set (Llama-3.3-70B was excluded because it was retired at rerun time), 11 context configurations, and 20 trial blocks per arm (4,400 model calls in total, none refused). Two independent implementations of the scoring pipeline agree on every value below. Because this rerun is an independent ten-model replicate rather than a reuse of the frozen data behind the +0.077 headline, its original-corpus arm (+0.09150) is a fresh measurement expected to differ from +0.077 through ordinary run-to-run variation.

The relevance-contrast gain (50:50 over 100% signal) stays clearly positive on both corpora, under both the primary and the matched-item scoring of the 100%-signal baseline:

| Scoring of the 100%-signal baseline | Original corpus | Corrected corpus | Corrected − original |
|---|---:|---:|---:|
| Over all target items (primary) | +0.09150 | +0.08325 | −0.00825 |
| Matched five items (sensitivity) | +0.08900 | +0.07050 | −0.01850 |

Both corrected-corpus points are positive and in the range of the paper's +0.077 headline, so the effect reproduces after cleaning. Both differences are negative — the corrected corpus shows a slightly smaller gain — which is the adverse (less-favorable) direction and is reported here in full rather than suppressed. The two rows are not independent. They share the 50:50 condition and differ only in how the 100%-signal baseline is scored, so they are one correlated pair rather than two separate confirmations.

The negative difference is not defensibly interpretable, for three reasons:

1. **Parse-failure sensitivity.** A small number of prompts returned output that could not be parsed (5 on the original corpus, 8 on the corrected). Holding out every such position symmetrically on both corpora and recomputing through the non-renormalizing scoring rule sends the gain — and therefore the difference — to *not estimable* on both arms. The negative sign does not survive a strict symmetric hold-out. A softer renormalizing check that keeps a usable number shrinks the difference by about three-quarters, from −0.00825 to −0.00202.
2. **Drift diagnostic.** The two arms were collected about ten hours apart, so a version-served model could in principle drift between them. Splitting the panel into the five version-pinned models (reproducible, no drift) and the five version-as-served models shows the difference is negative even in the version-pinned subset (−0.00250), so the gap between runs is not required to produce it. The version-as-served subset shows a larger negative value (−0.01400). These are small pooled diagnostics carried without uncertainty bands, so they place the difference inside the measurement-and-drift band rather than resolving its sign.
3. **Descriptive only.** The direction of the difference was not fixed in advance, so these points are descriptive comparisons rather than a directional test.

Taken together, the reproduction — both corpora clearly positive — is the substantive result. The small negative difference sits within measurement noise and run-to-run variation and is not read as a real reduction of the effect. A fuller robustness program, including independent re-rating with adjudication and rebuilt relevance pools, is left to future work.

**Per-model position sensitivity (exploratory).** The pooled C5 contrast above averages over all 11 model configurations, and the per-model estimates behind that average vary widely. Recomputed directly from the frozen per-trial results: DeepSeek-V3.2 +0.125 ($p=.0004$), GPT-5.4 +0.070 ($p=.049$), Grok-4-20 +0.045 ($p=.046$), Kimi-K2.6 +0.045 ($p=.046$), GPT-4o exactly flat, Llama-3.3-70B reversed at −0.065 ($p=.199$), and the remaining five configurations (Claude Opus 4.7 default/high/xhigh, Gemini-3.1-Pro-Preview, GPT-5.5) within ±0.03 of zero. These are single-model, unadjusted estimates — not corrected for the additional implicit comparisons across 11 models — reported as descriptive detail behind the pooled contrast, not as a second confirmatory result.

| Model configuration | End – start RelAcc | Raw *p* |
|---|---:|---:|
| DeepSeek-V3.2 | +0.125 | .0004 |
| GPT-5.4 | +0.070 | .049 |
| Grok-4-20 | +0.045 | .046 |
| Kimi-K2.6 | +0.045 | .046 |
| Claude Opus 4.7 xhigh | +0.025 | .367 |
| Gemini-3.1-Pro-Preview | +0.010 | .733 |
| GPT-5.5 | +0.005 | .874 |
| GPT-4o | 0.000 | 1.000 |
| Claude Opus 4.7 (default) | −0.005 | .924 |
| Claude Opus 4.7 high | −0.010 | .804 |
| Llama-3.3-70B | −0.065 | .199 |

## Practical Noise-Ratio Guidance

Recall keeps rising as the noise share increases, but relevance accuracy is highest across a broad plateau from roughly the 50:50 mix onward rather than at a single point — the practical operating range for extraction prompts. This connects the two confirmed contrasts. C1 (the 50:50-vs-100%-signal gain of +0.077) sits on that plateau, and C2 (the positive slope of +0.000871 RelAcc per percentage point of noise, fit on the design's nominal noise grid; the true item-count grid gives +0.000876) shows the gain accumulating gradually rather than in one step. Past roughly the halfway point, additional same-domain noise does not cost much relevance accuracy while continuing to help recall.

---

## Appendix E: Fusion-of-N Follow-Up

**Purpose:** Test whether learned synthesis of multiple extraction samples improves recall over simpler baselines.

**Setup:** DeepSeek-V3.2 generated five independent extraction samples per window. The same model then served as judge/fuser. The task was recall-oriented extraction over ecological dilution levels where noise and window length varied together.

The same model both generated the samples and served as fuser/judge. This self-preference bias would, if anything, favour the model's own fusion. The fact that a trivial set union still beat it makes the "union wins" result conservative rather than inflated.

**Result:** Learned fusion did not beat a free set union of extracted item IDs. The union reached the recall ceiling, while fusion sometimes dropped true positives found by only one sample. Recall here is item-identification recall — the fraction of target item IDs recovered — which is distinct from the main study's fact-type recall (Section 3.2).

| Comparison | Recall Difference | Interpretation |
|---|---:|---|
| Fusion - best-of-N | -0.048 | Reversed in the registered prompt, but prompt-dependent |
| Fusion - union(k=1) | -0.073 | Union is the stronger deployable baseline |
| Recall-tuned fusion - union(k=1) | -0.002 | Recall-tuned fusion ties, but does not beat, union |

**Conservative claim:** For recall-oriented extraction in this setup, use the set union of repeated samples before paying for learned fusion. Do not generalize this result to all aggregation tasks.